# An In-Depth Evaluation of Federated Learning on Biomedical Natural Language Processing

Le Peng[1], Gaoxiang Luo[2], Sicheng Zhou[3], Jiandong Chen[3], Ziyue Xu[4], Rui Zhang[5*], Ju Sun[1*]

**Affiliations:**

[1] Department of Computer Science and Engineering, University of Minnesota, Minneapolis, MN

[2] Department of Computer and Information Science, University of Pennsylvania, Philadelphia, PA

[3] Institute for Health Informatics, University of Minnesota, Minneapolis, MN

[4] Nvidia Corporation, Santa Clara, CA

[5] Division of Computational Health Sciences, Department of Surgery, University of Minnesota, Minneapolis, MN

*Corresponding authors:

Rui Zhang, Ph.D.

Division of Computational Health Sciences, Department of Surgery, University of Minnesota

Minneapolis, MN 55455

Email: zhan1386@umn.edu

Ju Sun, Ph.D.

Department of Computer Science and Engineering, University of Minnesota

Minneapolis, MN 55455

Email: jusun@umn.edu

# Abstract

Language models (LMs) such as BERT and GPT have revolutionized natural language processing (NLP). However, the medical field faces challenges in training LMs due to limited data access and privacy constraints imposed by regulations like the Health Insurance Portability and Accountability Act (HIPPA) and the General Data Protection Regulation (GDPR). Federated learning (FL) offers a decentralized solution that enables collaborative learning while ensuring data privacy. In this study, we evaluated FL on 2 biomedical NLP tasks encompassing 8 corpora using 6 LMs. Our results show that: 1) FL models consistently outperformed models trained on individual clients' data and sometimes performed comparably with models trained with polled data; 2) with the fixed number of total data, FL models training with more clients produced inferior performance but pre-trained transformer-based models exhibited great resilience. 3) FL models significantly outperformed large language models using zero-/one-shot learning and offered lightning inference speed.

# Introduction

The recent advances in deep learning have sparked the widespread adoption of language models (LMs), including prominent examples of BERT [1] and GPT [2], in the field of natural language processing (NLP). These LMs are trained on massive amounts of public text data, comprising billions of words, and have emerged as the dominant technology for various linguistic tasks, including text classification [3,4], text generation [5,6], information extraction [7–9], and question answering[10,11]. The success of LMs can be largely attributed to their ability to leverage large volumes of training data. However, in privacy-sensitive domains like medicine, data are often naturally distributed making it difficult to construct large corpora to train LMs. To tackle the challenge, the most common approach thus far has been to fine-tune pre-trained LMs for downstream tasks, using limited annotated data[12,13]. Nevertheless, pre-trained LMs are typically trained on text data collected from the general domain, which exhibits divergent patterns from that in the biomedical domain, resulting in a phenomenon known as domain shift. Compared to general text, biomedical texts can be highly specialized, containing domain-specific terminologies and abbreviations[14]. For example, medical records and drug descriptions often include specific terms that may not be present in general language corpora, and the terms often vary among different clinical institutes. Also, biomedical data lacks uniformity and standardization across sources, making it challenging to develop NLP models that can effectively handle different formats and structures. Electronic Health Records (EHRs) from different healthcare institutions, for instance, can have varying templates and coding systems[15]. So, direct transfer learning from LMs pre-trained on the general domain usually suffers a drop in performance and generalizability when applied to the medical domain as is also demonstrated in the literature[16]. Therefore, developing LMs that are specifically designed for the medical domain, using large volumes of domain-specific training data, is essential. Another vein of research explores pre-training the LM on biomedical data, e.g., BlueBERT[12], and PubMedBERT[17]. These LMs were either pre-trained on mixed-domain data (first pre-train on the general text and then keep pre-train on biomedical text) or directly pre-trained on domain-specific public medical datasets, e.g., PubMed literature and the Medical Information Mart for Intensive Care (MIMIC III)[18] and have shown improved performances compared to classical methods such as conditional random field (CRF)[19] and recurrent neural network (RNN) (e.g., long-short-term memory (LSTM)[20]) in many biomedical text mining tasks[8,9,12,16,21]. Nonetheless, it is important to highlight that the efficacy of these pre-trained medical LMs heavily relies on the availability of large volumes of task-relevant public data, which may not always be readily accessible.

All these mentioned above represent the classical *centralized learning* regime which involves aggregating data from distributed data sites and training a model in a single environment. However, this approach poses significant challenges in medicine, where data privacy is crucial, and data access is restricted due to regulatory concerns. Thus, in practice, people can only perform training with local datasets – *single-client training*. The drawback comes when the local dataset is small and often gives poor performance when evaluating an external dataset – poor generalization. To take advantage of the massively distributed data as well as improve the model generalizability, federated learning (FL) was initialized in 2016 [22] as a novel learning scheme to empower training with a decentralized environment and achieve many successes in critical domains with data privacy restrictions[23–25]. In an FL training loop, clients jointly train a shared global model by sharing the model weights or gradients while keeping their data stored locally. By bringing the model to the data, FL strictly ensures data privacy while achieving competitive levels of performance compared to a model trained with pooled data. While there is a rise of research showing great promise of applying FL in general NLP[26,27], applications of FL in biomedical NLP are still under-explored. Existing works in FL on biomedical NLP are either focused on optimizing one task[28,29] or trying to improve communication efficiency[28]. The current literature lacks a comprehensive comparison of FL on varied biomedical NLP tasks with real-world perturbations. To close this gap, we conducted an in-depth study of two representative NLP tasks, i.e., named entity recognition (NER) and relation extraction (RE), to evaluate the feasibility of adopting FL (e.g., FedAvg[30] and FedProx[31]) with LMs (e.g., Transformer-based models) in biomedical NLP. Our study aims to provide an in-depth investigation of FL in biomedical NLP by studying several FL variants on multiple practical learning scenarios including varied federation scales, different model architectures, data heterogeneities, and comparison with large language models (LLMs) on multiple benchmark datasets. Our major findings include:

1) When data were independent and identically distributed (IID), models trained using FL, especially pre-trained BERT-based models, performed comparable to centralized learning, a significant boost to single-client learning. Even when data were non-IID distributed, the gap can be filled by using alternative FL algorithms.

2) Larger models exhibited better resistance to the changes in FL scales. With a fixed number of data, the performance of FL models overall degraded as the clients' size increased. However, the deterioration diminished when combined with larger pre-trained models such as BERT-based models and GPT-2.

3) FL significantly outperformed large language models (LLMs), e.g., GPT-3, GPT-4, and PaLM 2, with zero-/one-shot learning, both in terms of prediction accuracy and inference speed.

# Results

In this section, we present our main results of analysis on FL with a focus on several practical facets, including 1) learning tasks, 2) scalability, 3) data distribution, 4) model architectures and sizes, and 5) comparative assessments with LLMs.

## FedAvg, Single-client, and Centralized learning for NER and RE tasks

Table 1 offers a summary of the performance evaluations for FedAvg, single-client learning, and centralized learning on five NER datasets, while Table 2 presents the results on three RE datasets. Our results on both tasks consistently demonstrate that FedAvg outperformed single-client learning. Notably, in cases involving large data volumes, such as BC4CHEMD and 2018 n2c2, FedAvg managed to attain performance levels on par with centralized learning, especially when combined with BERT-based pre-trained models.

**Table 1**. Comparison of FedAvg with centralized learning and single-client learning on 5 NER tasks measured by F1-score with lenient (upper) and strict (lower, inside parenthesis) matching scheme. For datasets involving multiple entities, we report the macro average score. The reported values represent the mean and standard deviation over three repeated experiments.

| Model | Method | 2018 n2c2 | BC2GM | BC4CHEMD | JNLPBA | NCBI-disease |
|---|---|---|---|---|---|---|
| BERT | Centralized | 0.879±0.002 (0.822±0.001) | 0.972±0.001 (0.928±0.001) | 0.981±0.001 (0.968±0.001) | 0.969±0.001 (0.939±0.002) | 0.989±0.001 (0.973±0.000) |
| | Single | 0.833±0.004 (0.766±0.002) | 0.886±0.001 (0.755±0.002) | 0.924±0.001 (0.883±0.000) | 0.905±0.001 (0.813±0.002) | 0.918±0.003 (0.842±0.003) |
| | FedAvg | 0.877±0.002 (0.817±0.002) | 0.959±0.001 (0.897±0.000) | 0.973±0.000 (0.954±0.001) | 0.949±0.001 (0.896±0.001) | 0.976±0.001 (0.949±0.001) |
| BlueBERT | Centralized | 0.879±0.005 (0.820±0.007) | 0.975±0.000 (0.932±0.002) | 0.965±0.004 (0.944±0.007) | 0.969±0.001 (0.940±0.003) | 0.987±0.008 (0.968±0.009) |
| | Single | 0.836±0.004 (0.767±0.005) | 0.904±0.003 (0.775±0.003) | 0.930±0.001 (0.895±0.003) | 0.907±0.001 (0.817±0.003) | 0.929±0.004 (0.857±0.006) |
| | FedAvg | 0.876±0.002 (0.817±0.000) | 0.966±0.001 (0.919±0.002) | 0.977±0.000 (0.959±0.000) | 0.963±0.001 (0.923±0.001) | 0.984±0.002 (0.963±0.000) |
| BiLSTM-CRF | Centralized | 0.834±0.002 (0.783±0.002) | 0.924±0.001 (0.866±0.001) | 0.958±0.001 (0.934±0.001) | 0.961±0.000 (0.924±0.001) | 0.971±0.002 (0.944±0.004) |
| | Single | 0.734±0.001 (0.667±0.006) | 0.619±0.005 (0.409±0.014) | 0.764±0.002 (0.669±0.007) | 0.824±0.003 (0.669±0.010) | 0.738±0.012 (0.589±0.041) |
| | FedAvg | 0.782±0.002 (0.734±0.003) | 0.793±0.005 (0.645±0.013) | 0.920±0.002 (0.882±0.002) | 0.902±0.001 (0.810±0.004) | 0.865±0.020 (0.767±0.035) |
| BioBERT | Centralized | 0.884±0.002 (0.823±0.002) | 0.980±0.000 (0.937±0.003) | 0.983±0.001 (0.972±0.001) | 0.971±0.000 (0.943±0.001) | 0.993±0.001 (0.975±0.001) |
| | Single | 0.849±0.004 (0.784±0.003) | 0.927±0.001 (0.808±0.001) | 0.945±0.001 (0.913±0.001) | 0.917±0.001 (0.828±0.002) | 0.937±0.001 (0.870±0.008) |
| | FedAvg | 0.879±0.002 (0.818±0.003) | 0.974±0.001 (0.922±0.000) | 0.978±0.000 (0.963±0.001) | 0.957±0.001 (0.910±0.002) | 0.983±0.002 (0.958±0.001) |
| Bio_clincialBERT | Centralized | 0.885±0.006 (0.827±0.005) | 0.974±0.001 (0.933±0.001) | 0.980±0.001 (0.967±0.001) | 0.969±0.001 (0.941±0.001) | 0.993±0.001 (0.975±0.001) |
| | Single | 0.847±0.002 (0.782±0.002) | 0.892±0.001 (0.765±0.004) | 0.925±0.002 (0.885±0.003) | 0.904±0.001 (0.815±0.001) | 0.927±0.001 (0.854±0.008) |
| | FedAvg | 0.878±0.001 (0.815±0.001) | 0.960±0.002 (0.901±0.001) | 0.971±0.001 (0.953±0.001) | 0.951±0.000 (0.901±0.001) | 0.982±0.003 (0.958±0.004) |
| GPT-2 | Centralized | 0.801±0.001 (0.745±0.001) | 0.891±0.001 (0.836±0.001) | 0.879±0.002 (0.857±0.002) | 0.925±0.001 (0.881±0.001) | 0.928±0.002 (0.904±0.002) |
| | Single | 0.741±0.005 (0.685±0.007) | 0.708±0.010 (0.549±0.011) | 0.747±0.004 (0.687±0.006) | 0.793±0.004 (0.669±0.005) | 0.765±0.012 (0.684±0.013) |
| | FedAvg | 0.798±0.003 (0.746±0.001) | 0.796±0.001 (0.674±0.006) | 0.825±0.000 (0.794±0.000) | 0.844±0.001 (0.748±0.001) | 0.852±0.003 (0.809±0.002) |

**Table 2.** Comparison of FedAvg with centralized learning and single-client learning on RE task measure by macro F1-score. The reported values represent the mean and standard deviation over three repeated experiments.

| Model | Method | 2018 n2c2 | EUADR | GAD |
|---|---|---|---|---|
| BERT | Centralized | 0.947±0.001 | 0.750±0.040 | 0.738±0.028 |
| | Single | 0.887±0.008 | 0.576±0.154 | 0.652±0.010 |
| | FedAvg | 0.946±0.002 | 0.527±0.008 | 0.703±0.021 |
| BlueBERT | Centralized | 0.950±0.002 | 0.582±0.109 | 0.755±0.007 |
| | Single | 0.896±0.010 | 0.420±0.048 | 0.663±0.021 |
| | FedAvg | 0.950±0.002 | 0.548±0.073 | 0.714±0.018 |
| BioBERT | Centralized | 0.942±0.002 | 0.737±0.049 | 0.783±0.007 |
| | Single | 0.895±0.009 | 0.640±0.119 | 0.672±0.015 |
| | FedAvg | 0.942±0.002 | 0.718±0.037 | 0.750±0.008 |
| Bio_ClincialBERT | Centralized | 0.950±0.001 | 0.741±0.067 | 0.743±0.014 |
| | Single | 0.902±0.004 | 0.620±0.138 | 0.589±0.034 |
| | FedAvg | 0.946±0.003 | 0.578±0.057 | 0.695±0.009 |
| GPT-2 | Centralized | 0.951±0.004 | 0.684±0.097 | 0.709±0.004 |
| | Single | 0.893±0.013 | 0.279±0.104 | 0.630±0.008 |
| | FedAvg | 0.946±0.003 | 0.547±0.086 | 0.721±0.009 |

Influence of FL scale on the performance of LMs

In clinical applications, there are two distinct learning paradigms. The first involves small-scale client cohorts, each equipped with substantial data resources, often seen in collaborations within hospital networks. In contrast, the second encompasses widely distributed clients, characterized by more limited data holders, often associated with collaborations within clinical facilities or on mobile platforms. We investigated the performance of FL on the two learning paradigms by varying client group sizes while maintaining a fixed total training data volume. The results are summarized in Fig. 1, revealing a consistent trend: notably larger models, such as those backed by BERT and GPT-2 architectures, exhibited great resilience to fluctuations in federation scales. In contrast, the lightweight model, as of BiLSMT-CRF, was susceptible to alterations of scale, resulting in a rapid deterioration in performance as the number of participating clients increased.

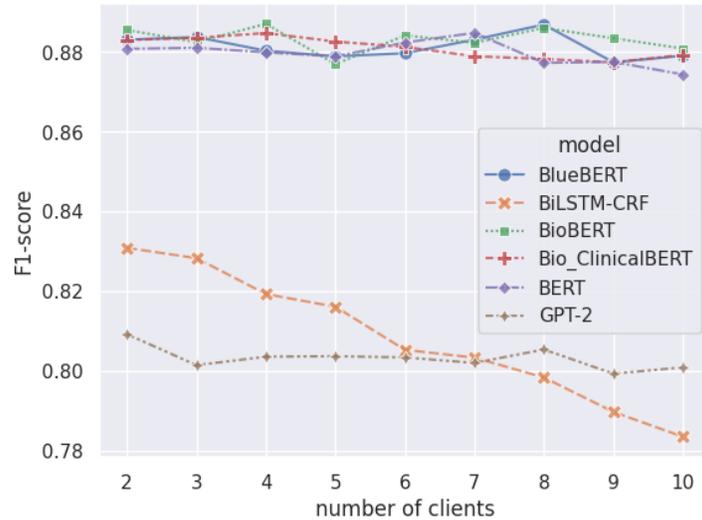

**Fig. 1** Performance of FL models with varying numbers of clients

## Comparison of FedAvg and FedProx with data heterogeneity

Biomedical texts often exhibit high specialization due to distinct protocols employed by different hospitals when generating medical records, resulting in great variations — sublanguage differences. Therefore, FL practitioners should account for such data heterogeneity when implementing FL in healthcare systems. We simulated a real non-IID scenario by emulating BC2GM and JNLPBA as two clients and jointly performing FL. We considered two FL algorithms including FedAvg and FedProx, both are widely deployed in practice. For comparison, we also studied a simulated IID setting using the 2018 n2c2 dataset by random splitting. As shown in Table 3, we observed that the performance of FedProx was sensitive to the choice of the hyper-parameter $\mu$. Notably, a smaller $\mu$ consistently resulted in improved performance. When $\mu$ was carefully selected, FedProx outperformed FedAvg when the data were non-IID distributed (lenient F1 score of 0.994 vs. 0.934, and strict F1 score of 0.901 vs. 0.884). However, the difference between the two algorithms was mostly indistinguishable when the data were IID distributed (lenient F1 score of 0.880 vs. 0.879, and strict F1 score of 0.820 vs. 0.818).

**Tabl**e 3. Comparison of FedAvg with centralized learning and single-client learning using bioBERT. We select the value of $\mu$ (a hyper-parameter in FedProx) as suggested by the FedProx paper. The reported values represent the mean and standard deviation over three repeated experiments.

| Method | $\mu$ | IID (2018 n2c2) | | non-IID (BC2GM & JNLPBAS) | |
|---|---|---|---|---|---|
| | | lenient | strict | lenient | strict |
| Centralized | - | 0.884±0.002 | 0.823±0.002 | 0.964±0.001 | 0.929±0.000 |
| FedAvg | - | 0.879±0.002 | 0.818±0.003 | 0.934±0.003 | 0.884±0.003 |
| FedProx | 1 | 0.855±0.003 | 0.790±0.005 | 0.880±0.001 | 0.772±0.002 |
| | 0.5 | 0.868±0.001 | 0.809±0.002 | 0.881±0.002 | 0.777±0.001 |
| | 0.1 | 0.872±0.003 | 0.814±0.004 | 0.897±0.002 | 0.817±0.002 |
| | 0.01 | 0.878±0.003 | 0.819±0.002 | 0.933±0.002 | 0.884±0.003 |
| | 0.001 | 0.880±0.002 | 0.820±0.001 | 0.944±0.002 | 0.901±0.002 |

## Impact of the LM size on the performance of different training schemes

We investigated the impact of model size on the performance of FL. We compared 6 models with varying sizes with the smallest one comprising 20 M parameters and the largest one comprising 334M parameters. We picked the BC2GM dataset for illustration and anticipated similar trends would hold for other datasets as well. As shown in Fig. 2, in most cases, larger models (represented by large circles) overall exhibited better test performance than their smaller counterparts. For example, BlueBERT demonstrated uniform enhancements in performance compared to BiLSTM-CRF and GPT2. Among all the models, BioBER emerged as the top performer, whereas GPT-2 gave the worst performance.

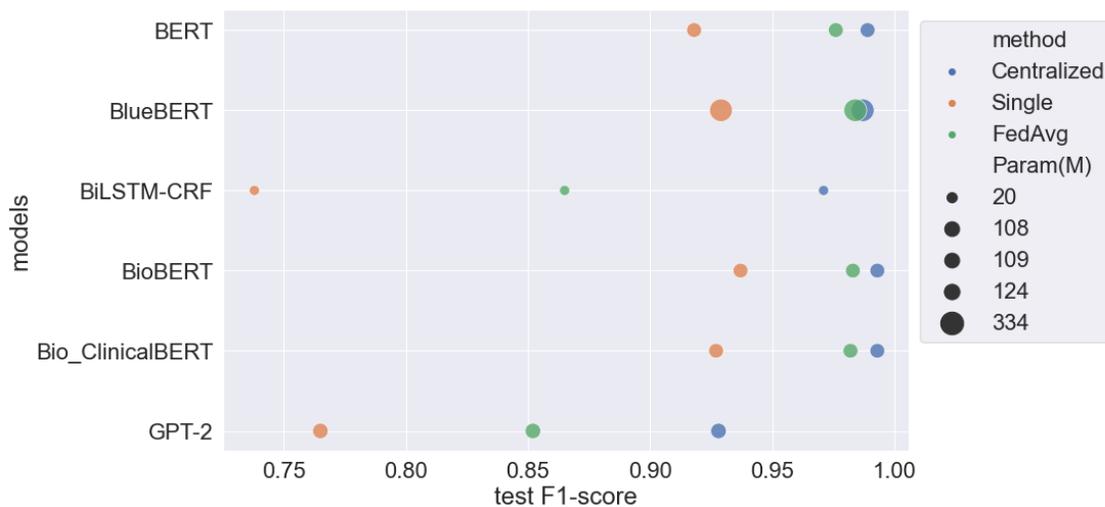

**Fig. 2** Comparison of model performance with different sizes, measured by the number of trainable parameters on the BC2GM dataset. The size of the circle tells the number of model parameters, while the color indicates different learning methods. The x-axis represents the mean test F1-score with the lenient match.

## Comparison between FL and LLM

**Table 4.** Comparison of LLM and FL on NER and RE tasks. We report strict and lenient (in parenthesis) F1-score for NER, and macro F1-score for RE. For both tasks, the inference speed is measured by the time to process one instance (sentence).

| Model | Method | NER | | | RE | | |
|---|---|---|---|---|---|---|---|
| | | 2018 n2c2 | NCBI-disease | instance/s | 2018 n2c2 | GAD | instance/s |
| GPT-3 | zero-shot | 0.294 (0.446) | 0.382 (0.564) | 7.38 | 0.204 | 0.451 | 0.47 |
| GPT-3 | one-shot | 0.493 (0.621) | 0.374 (0.604) | 7.63 | 0.294 | 0.462 | 0.46 |
| GPT-4 | zero-shot | 0.486 (0.701) | 0.602 (0.773) | 5.90 | 0.638 | 0.494 | 0.86 |
| GPT-4 | one-shot | 0.636 (0.774) | 0.621 (0.764) | 6.63 | 0.739 | 0.459 | 0.76 |
| PaLM 2 | zero-shot | 0.226 (0.404) | 0.552 (0.676) | 1.05 | 0.574 | 0.600 | 0.52 |
| PaLM 2 | one-shot | 0.470 (0.591) | 0.527 (0.644) | 1.03 | 0.618 | 0.471 | 0.55 |
| **BlueBERT** | **FL** | **0.824 (0.899)** | **0.954 (0.986)** | **1.49x10$^{-3}$** | **0.969** | **0.742** | **2.61x10$^{-2}$** |
| **GPT-2** | **FL** | **0.784 (0.840)** | **0.830 (0.868)** | **6.12x10$^{-4}$** | **0.946** | **0.721** | **1.58x10$^{-2}$** |

In light of the well-demonstrated performance of large language models (LLMs) on various linguistic tasks, we explored the performance gap of LLMs to the smaller LMs trained using FL. Notably, it is usually not common to fine-tune LLMs due to the formidable computational costs and protracted training time. Therefore, we selected two representative methods that enable direct inference from pre-trained LLMs, specifically zero-shot and one-shot learning, and compared them with models trained using FL. We followed the experimental protocol outlined in a recent study[32] and evaluated all the models on two NER datasets (2018 n2c2 and NCBI-disease) and two RE datasets (2018 n2c2, and GAD). The results, as summarized in Table 4, underscore that FL, whether implemented with a BERT-based model or GPT-2 model, consistently outperformed GPT-3 and even surpassed GPT-4 and PaLM 2 with both zero-shot and one-shot learning. Beyond the performance gains, FL trained with small LMs also offered substantially faster inference speeds.

## Discussion

In this study, we visited FL for biomedical NLP and studied two established tasks (NER and RE) across 7 benchmark datasets. We examined 6 LMs with varying parameter sizes (ranging from BiLSTM-CRF with 20 M to transformer-based models up to 334 M parameters) and compared their performance using centralized learning, single-client learning, and federated learning. On almost all the tasks, we showed that federated learning achieved significant improvement compared to single-client learning, and oftentimes performed comparably to centralized learning without data sharing, demonstrating it as an effective approach for privacy-preserved learning with distributed data. The only exception is in Fig. 8, where single-client learning outperformed FedAvg when using BERT and bio_ClinicalBERT. We believe this is due to the lack of training data. As each client only owned 28 training sentences, the data distribution, although IID, was highly under-represented, making it hard for FedAvg to find the global optimal solutions. Surprisingly, FL achieved reasonably good performance even when the training data was limited (284 total training sentences from all clients), confirming that transfer learning from either the general text domain (e.g.,

BERT and GPT-2) or biomedical text domain (e.g., blueBERT, bioBERT, bio_ClinicalBERT) is beneficial to the downstream biomedical NLP task and pretraining on medical data often gives a further boost. Another interesting finding is that GPT-2 always gave inferior results compared to BERT-based models. We believe this is because GPT-2 is pre-trained on text generation tasks that only encode left-to-right attention for the next word prediction. However, this unidirectional nature prevents it from learning more about global context which limits its ability to capture dependencies between words in a sentence.

In the sensitivity analysis of FL to client sizes, we found there is a monotonic trend that, with a fixed number of training data, FL with fewer clients tends to perform better. For example, the classical BiLSTM-CRF model (20M), with a fixed number of total training data, performs better with few clients, but performance deteriorates when more clients join in. It is likely due to the increased learning complexity as FL models need to learn the inter-correlation of data across clients. Interestingly, the transformer-based model (>= 108M), which is over 5 sizes larger compared to BILSMT-CRF, is more resilient to the change of federation scale, possibly owing to its increased learning capacity.

We analyzed the performance of FedProx in real-world non-IID scenarios and compared it with FedAvg to study the behavior of different FL algorithms under data heterogeneity. Although the FedProx achieved slightly better performance than FedAvg when the data were non-IID distributed, it is very sensitive to the hyper-parameter $\mu$ which strikes to balance the local objective function and the proximal term. Specifically, when data was IID and $\mu$ was set to a large value (e.g., $\mu$=1), FedProx yielded a 2.4% lower lenient F1-score compared to FedAvg. When the data were non-IID, this performance gap further widened to 5.4%. It is also noteworthy that when $\mu$ is set to 0, and all the clients are forced to perform an equal number of local updates, FedProx essentially reverts to FedAvg.

We also investigated the impact of model size on the performance of FL. We observed that as the model size increased, the performance gap between centralized models and FL models narrowed. Interstingly, BioBERT, which shares the same model architecture and is similar in size to BERT and Bio_ClinicalBERT, performs comparably to larger models (such as BlueBERT), highlighting the importance of pre-training for model performance. Overall, the size of the model is indicative of its learning capacity,  large models tend to perform better than smaller ones. However, large models require longer training time and more computation resources which results in a natural trade-off between accuracy and efficiency.

In comparison with LLM, FL models were the winner both in terms of prediction accuracy and inference speed. We hypothesize that LLMs, although perform well on general linguistic tasks, can not easily adapt to the specialized tasks given zero/one sample as input. To close the gap and make better use of LLMs given the context of biomedical NLP, specialized LLMs that are pre-trained on medical text data [33] or model fine-tuning [34] are needed.

While seeing many promising results of FL for LMs, we acknowledge our study suffers from the following limitations: 1) most of our experiments, excluding the non-IID study, are conducted in a simulated environment with synthetic data split, which may not perfectly align with the distribution patterns of real-world FL data. 2) we mostly focused on horizontal FL, but have not extended to vertical FL[35]. 3) we have not considered FL combined with privacy techniques such as differential privacy[36] and homographic encryption[37]. To address these limitations and further advance our understanding of FL for LMs, our future study will focus on the real-world implementation of FL and explore the practical opportunities and challenges in FL such as vertical FL and FL combined privacy techniques. We believe our study will offer comprehensive insights into the potential of FL for LMs, which can serve as a catalyst for future research to develop more effective AI systems by leveraging distributed clinical data in real-world scenarios.

## Methods

### NLP tasks and corpora

We compared FL with alternative training schemes on 8 biomedical NLP datasets with a focus on two NLP tasks: NER (5 corpora) and RE (3 corpora). The NER and RE are two established tasks for information extraction in biomedical NLP. Given an input sequence of tokens, the goal of NER is to identify and classify the named entities, such as diseases and genes, present in the sequence. RE is often the follow-up task that aims to discover the relations between pairs of named entities. For example, a gene-disease relation (BRCA1-breast cancer) can be identified in a sentence "Mutations of BRCA1 gene are associated with breast cancer". For RE tasks, we take the entity positions as given and formulate the problem as follows: given a sentence and the spans of two entities, the task is to determine the relationship between the two entities.

**Table 5.** List of corpora and their statistics. The data splits are counted based on the number of sentences.

| Corpus | Entity/Relation Type | Task | Train | Dev | Test |
| --- | --- | --- | --- | --- | --- |
| 2018 n2c2[38] | 8 entities[1] | NER | 48727 | 6091 | 6091 |
| BC2GM[39] | gene | NER | 26006 | 3251 | 3251 |
| BC4CHEMD[40] | drug/chem | NER | 94170 | 11772 | 11771 |
| JNLPBA[41] | gene | NER | 29559 | 3695 | 3695 |
| NCBI-disease[42] | disease | NER | 10125 | 1266 | 1266 |
| 2018 n2c2[38] | disease | RE | 72786 | 9099 | 9098 |
| EUADR[43] | gene-disease | RE | 284 | 36 | 35 |
| GAD[21] | gene-disease | RE | 4097 | 513 | 512 |

For all NER corpora, it follows the same BIO notation to distinguish the beginning (B), inside (I), and outside (O) of entities. We adopted most of the preprocessed corpora from the paper of BioBERT[8], except for the 2018 n2c2 dataset (both NER and RE). For all the datasets, we removed duplicated notes and split the data into the train(80%), dev(10%), and test(10%). A summary of the datasets can be found in Table 5, we defer to supplementary materials for more detailed descriptions for each dataset.

---

[1] A total of 9 entities are considered including *reason, frequency, ADE, strength, duration, route, form, and dosage*. Details about the 2018 n2c2 dataset can be found in supplementary materials.

## Federated learning algorithms

FL represents a family of algorithms that aims to train models in a distributed environment in a collaborative manner. Consider a scenario where there are K clients with distributed data $D = \{D_1, D_2, ..., D_k\}$, where $D_i = D_{X_i \times Y_i}$, and $X_i$ and $Y_i$ are the input and output space, respectively. The typical FL aims to solve the optimization problem:

$$\sum_{i=1}^{K} p_i F_i(w) \quad \text{where } F_k = \sum_{j=1}^{|D_k|} L_w(X_j, Y_j),$$

where $w$ denote the weights of the model being learned, $F_i$ is the local objectives of $i$-th clients and $p_i$ is the weights of the $i$-th clients such that $p_i > 0$ and $\sum_{i=1}^{K} p_i = 1$. The weights are usually determined by the quantity of clients' training samples. For example, it equals $\frac{1}{K}$ when clients share the same amount of training data.

In an FL game, there are two types of players: server and client. The server is the compass that navigates the whole process of FL including signaling the start and end of federated learning, synchronizing the local model updates, and dispatching the updated models. The clients are responsible for fetching models from the server, updating models using their local data, and sending the updated models back to the server.

Throughout the whole process, there are two major steps: 1) the clients use their own data to optimize the local objectives — **local updates**, 2) local clients upload the updated model or gradients to the server, 3) the server acquires the local models and synchronize the updates — **model aggregation**, and 4) server dispatch the models to the clients. While different FL algorithms may have specialized designs for local updates or model aggregation, they share the same training paradigm.

We considered the two most popular FL algorithms called Federated Averaging (FedAvg)[30] and another variant FedProx[31]. **FedAvg** is the most basic and standard FL algorithm that uses stochastic gradient descent (SGD) to progressively update the local model. More specifically, each client locally takes a fixed number of gradient descent steps on their local model using their local training data. On another hand, the server will aggregate these local models by taking the weighted average as the resulting new model for the next round. However, in FedAvg, the number of local updates can be determined by the size of the data. When the size of the data varies, the local updates performed locally can be significantly different. **FedProx** was introduced to tackle the issue of heterogeneous local updates in FedAvg. By adding a proximal term to the objective of the local update, the impact of variable local updates is suppressed. More specifically, at iteration t, the inner local updates are trying to find the solution that minimizes the following objective

$$Min_w \frac{1}{n_k} \sum_{i=1}^{n_k} L_w(X_i, Y_i) + \frac{\mu}{2} ||w - w^t||$$

where $w^t$ is the weights of the network from the iteration t. A comparison of FedAvg and FedProx can be found in Algorithm 1 and Algorithm 2 in supplementary materials.

Study design

As shown in Fig. 2, we explored three learning methods: 1) federated learning, centralized learning, and single-client learning. To simulate the conventional learning scenario, we varied the data scale and conducted the following experiments: centralizing all client data to train a single model (centralized learning) and training separate models on each client's local data (single-client learning).

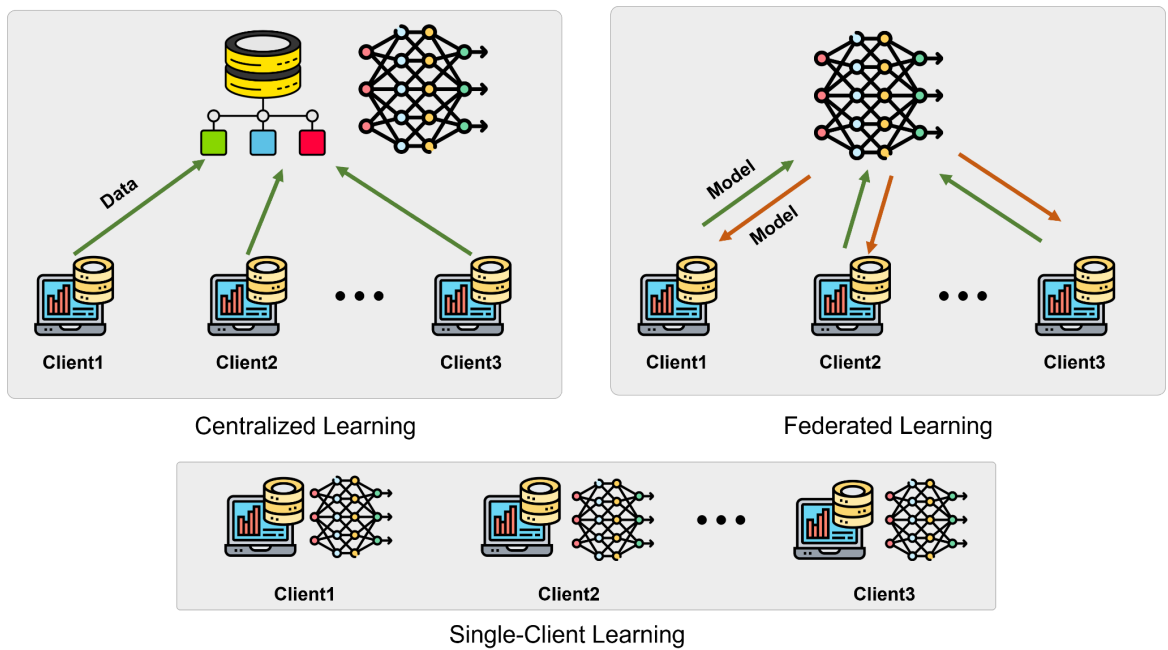

**Fig. 2** A comparison of centralized learning, federated learning, and single-client learning.

**Models:** To better understand the effect of LMs on FL, we chose models with various sizes of parameters from 20 M to 334 M including Bidirectional Encoder Representations from Transformer(BERT)[1], and Generative Pre-trained Transformer (GPT), as well as classical RNN-based model like BiLSTM-CRF[44]. BERT-based models utilize a transformer encoder and incorporate bi-directional information acquired through two unsupervised tasks as a pre-training step into its encoder. Different BERT models differ in their pre-training source dataset and model size, deriving many variants such as BlueBERT[12], BioBERT[8], and Bio_ClinicBERT[45]. BiLSTM-CRF is the only model in our study that is not built upon transformers. It is a bi-directional model designed to handle long-term dependencies, is used to be popular for NER, and uses LSTM as its backbone. We selected this model in the interest of investigating the effect of federation learning on models with smaller sets of parameters. For LLMs, we selected GPT-3, GPT-4, and PaLM 2 for assessment as both can be publicly accessible for inference. A summary of the model can be found in Table 6, and details on the model description can be found in the supplementary materials.

**Table 6.** List of LMs used for comparison.

| Model | Param | Backbone | pre-trained Source | Year |
|---|---|---|---|---|
| BiLSTM-CRF[44] | 20M | LSTM | - | 2015 |
| BERT[1] | 109M | Transformer encoder | Wikipedia + BooksCorpus | 2018 |
| BlueBERT[12] | 334M | Transformer encoder | PubMed | 2019 |
| BioBERT[8] | 108M | Transformer encoder | Wikipedia + BooksCorpus + PubMed + PMC | 2020 |
| Bio_ClinicalBERT[9] | 108M | Transformer encoder | clinical notes | 2019 |
| GPT-2[46] | 124M | Transformer decoder | Wikipedia +news+books | 2019 |
| GPT-3[47] | 175B | Transformer decoder | CommonCrawl+BooksCorpus+Wikipedia+WebText2 | 2020 |
| GPT-4[48] | - | Transformer decoder | - | 2023 |
| PaLM 2[49] | - | Transformer | web documents, books, code, mathematics, and conversational data | 2023 |

## Training details

**Data Preprocessing**: we adapted most of the dataset from the BioBERT paper with reasonable modifications by removing the duplicate entries and redoing the data splits; details of cleaning steps can be found in the supplementary materials. The maximum token limit was set at 512, with truncation—coded sentences with length larger than 512 were trimmed.

**Federated learning simulation**: We considered two different learning settings: learning from independent and identically distributed (IID) data and learning from non-IID data. For the first setting, we randomly split the data into k folds uniformly. For the majority of our experiments, k was chosen as 10, while we also varied k from 2 to 10 to study the impact of the size of the federation. For the second setting, we considered learning from heterogeneous data collected from different sources. This represents the real-world scenario where complex and entangled heterogeneities are co-existed. We picked the BC2GM and JNLPBA as two independent clients, both are targeting the same gene entity recognition tasks but were collected from different sources.

**LLMs with zero-/one-shot learning:** We followed the experiment protocol as in the previous study[32]. In NER, the prompts for zero-shot are designed as:

*"**Task**: the task is to extract disease entities in a sentence"*

*"**Input**": the input is a sentence."*

*"**Output**: the output is an HTML that highlights all the disease entities in the sentence. The highlighting should only use HTML tags <span style=\"background-color: #FFFF00\"> and </span> and no other tags."*

For one-shot, we add an example of input and expected outputs:

*"**Example**:*

***Input**: In summary, inactivation of the murine ATP7B gene produces a form of cirrhotic liver disease that resembles Wilson disease in humans and toxic milk phenotype in the mouse"*

***Output**: In summary, inactivation of the murine ATP7B gene produces a form of <span style="background-color: #FFFF00> cirrhotic liver disease</span> that resembles <span style="background-color: #FFFF00>Wilson disease</span> in humans and toxic milk phenotype in the mouse"*

For model evaluation, we randomly selected 200 test samples in the test dataset and reported the prediction performance over the selected samples.

**Training Models**: We used Adam to optimize our models with an initial learning rate of 0.001 and momentum of 0.9. The learning rate was scheduled by *linear_scheduler_with_warmup*. All experiments were performed on a system equipped with an NVIDIA A100 GPU and an AMD EPYC 7763 64-core Processor.

**Reported evaluation:** For NER, we reported the performance of these metrics at the macro average level with both strict and lenient match criteria. Strict match considers the true positive when the boundary of entities exactly matches with the gold standard, while lenient considers true positives when the boundary of entities overlaps between model outputs and the gold standard. For all tasks, we repeated the experiments three times and reported the mean and standard deviation to account for randomness.

## Data Availability

All the datasets involved in this study are publicly available from the following official websites:

2018 n2c2: https://portal.dbmi.hms.harvard.edu/projects/n2c2-nlp/

BC2GM: https://biocreative.bioinformatics.udel.edu/tasks/

BC4CHEMD: https://biocreative.bioinformatics.udel.edu/resources/biocreative-iv/chemdner-corpus/

JNLPBA: http://www.geniaproject.org/shared-tasks/bionlp-jnlpba-shared-task-2004

NCBI-disease: https://www.ncbi.nlm.nih.gov/CBBresearch/Dogan/DISEASE/

EUADR: https://biosemantics.erasmusmc.nl/index.php/resources/euadr-corpus

GAD: https://maayanlab.cloud/Harmonizome/dataset/GAD+Gene-Disease+Associations

## Code Availability

Our project codes are publicly available on Github:

Train and evaluate FL models: https://github.com/PL97/FedNLP

Texts preprocessing: https://github.com/PL97/Brat2BIO

Evaluation: https://github.com/PL97/NER_eval

LLMs evaluations: https://github.com/GaoxiangLuo/LLM-BioMed-NER-ER

## Acknowledgments

This work was in part supported by Cisco Research under award number 1085646 PO USA000EP390223. The authors acknowledge the Minnesota Supercomputing Institute (MSI) at the University of Minnesota for providing resources that contributed to the research results reported in this paper.

## Author Contributions

L.P. was responsible for the overall experimental design, FL implementation, and writing of the manuscript. G.L. was responsible for the LLM prompt design, LLM experiment, evaluation, and editing of the manuscript. S.Z. and R.Z. contributed to the data collection and editing of the manuscript. J.C., Z.X, and J.S. contributed to the editing of the manuscript and idea discussion.

# Supplementary Materials

Experiment details

**Data Preprocessing**: We removed duplicated notes and split the data into the non-overlapped train(80%), dev(10%), and test(10%) datasets. To simulate an FL setting, we further split the train and dev set equally into 2/5/10 folds to mimic varied numbers of clients' participation.

**Hyperparameter tuning**: For models involving one or more hyper-parameters, we applied grid search to find the best combination of hyper-parameters.

**FL Algorithms:** Considering there are numerous FL algorithms designed to tackle various problems, we pick two fundamental algorithms, FedAvg and FedProx, both are the most representative and popular FL algorithms that are widely used in practice. We showed the comparisons between the two algorithms as detailed in Algorithm 1 (server aggregation) and Algorithm 2 (clients' local updates).

---

**Algorithm 1**: Federated learning algorithms (FedAvg/FedProx)

---

Notation: $X_i$ indicate data from client i, K is the total number of clients, T is maximum training round, n is the sum of $n_1$ to $n_k$, σ is the hyper-parameter in FedAMP

Initialize server model weights $\mathbf{w}(1)$
Initialize client model weights $w_i \ \forall \ i = 1, 2, ..., K$
For each round t = 1, 2, … T do
    Send server model weight $w(t)$ to each client
    For each client $k = 1, 2, ..., K$ do
        Client k perform LocalUpdate($X_k, Y_k, w_k$)     ← Algorithm 2

$$\gamma_k = \frac{n_k}{n}$$
    End for
End for

---

**Algorithm 2**: Local model training using mini-batch stochastic gradient descent (LocalUpdate) (FedAvg/FedProx)

---

Notation: R is the local update round, B is the number of batches, $f_{w_r}$ is the neural network parameterized by $w_r$, η is the learning rate, μ is the hyper-parameter in FedProx, λ and $α_k$ is the hyper-parameters in FedAMP

For each round $r = 1, 2, ..., R$ do / Repeat until find the approximate minimizer of
$w ≈ argmin_w L(f_{w_r}(X_b), Y_b) + \frac{\mu}{2}||w_k - w_k(t)||^2$
    Randomly shuffle $X_k$ and create B batches $((X_1, Y_1), (X_2, Y_2), ..., (X_B, Y_B))$
    $L_{w_r} = L(f_{w_r}(X_b), Y_b) + \frac{\mu}{2}||w_k - w_k(t)||^2$
    For each mini-batch $b = 1, 2, ..., B$ do
        $w_{r+1} = w_r - η∇L_{w_r}(X_b, Y_b)$
End for

---

## Datasets

**2018 National NLP Clinical Challenges (n2c2) Shared Task[38]**: 2018 n2c2 corpus contains 505 discharge summaries from the MIMIC-III clinical care database[4]. The goal of the task is to extract entity tags (*reason, frequency, ADE, strength, duration, route, form, drug, and dosage*) that indicate the presence of drug and ADE information, and relations (*strength-drug, duration-drug, route-drug, form-drug, ADE-drug, Dosage-drug, reason-drug, and frequency-drug*) between the entities.

**BioCreative II Gene Mention Recognition (BC2GM)**[39]: BC2GM Dataset collected text data related to gene information. The dataset comprises a set of sentences, and a set of gene mentions (GENE annotations) for each sentence. Some GENE annotations in a sentence may also have alternate boundaries that are judged by human annotators which can be essentially equivalent references (ALTGENE annotations). The goal of the task is to identify gene mentions in a sentence according to its start and end characters.

**BioCreative IV Chemical Compound and Drug Name Recognition (BC4CHEMD)**[40]: BC4CHEMD contains a total of 84,355 chemical mention annotations from 10,000 PubMed abstracts which are manually labeled by some chemistry literature experts. The goal of the task is to classify the text into multiple CEM classes: *systematic, identifiers, formula, trivial, abbreviation, family, and multiple.*

**JNLPBA**[41]: JNLPBA originated from the GENIA version 3.02. It is a selection of 2,000 abstracts with a controlled search on MEDLINE using the MeSH terms *'human'*, *'blood cells'*, and *'transcription factors'*. The abstracts were hand-annotated to 36 terminal classes according to a small taxonomy of 48 classes based on a chemical classification.

**NCBI-disease**[42]: The NCBI-disease corpus is collected from 793 PubMed abstracts that are fully annotated at the disease mentions and concept level based on corresponding identifiers from either Medical Subject Headings (MeSH) or Online Mendelian Inheritance in Man (OMIM). It includes 6892 disease mentions, which are mapped to 790 unique disease concepts. 12% link to an OMIM identifier, while the remaining contain a MeSH identifier. In addition, 91% of mentions are described as a single disease concept, while the remaining link to a combination of concepts.

**EUADR**[43]: EUADR corpus was annotated for disorders, drugs, genes, and their inter-relationships. Three experts were used to annotate a set of 100 abstracts for each of the drug-disorder, drug-target, and target-disorder relations. The drug-disorder and drug-target relations were composed of 100 randomly selected abstracts from the PubMed result. For the target-disorder set, 50 abstracts were randomly selected from gene disorder, and 50 abstracts were randomly selected from SNP-disorder relation.

**Gene Associations Database (GAD)**[21]: Gene Associations Dataset is a corpus that provides a public, comprehensive repository of molecular, clinical, and study parameters for > 5000 human genetic association studies to explore gene-disease relations. It contains 10697 genes, 12774 diseases, and 74928 gene-disease associations.

## Models Architectures

**BERT**[1]: Bidirectional Encoder Representations from Transformer (BERT) was developed in 2018 by researchers at Google. The BERT's model architecture is a multi-layer bidirectional Transformer encoder. It employs encoders as a sub-structure to pre-training models for NLP tasks. BERT comprehends language via Masked Language Modeling (MLM) and Next Sentence Prediction (NSP) mechanisms. By assuming a blinder with MLM, BERT learns bidirectional contexts within sentences. The model takes random sentences as input, masks certain words, and then reconstructs the masked words from the surrounding text. BERT's ability to process two sentences simultaneously and determine if the second follows the first enables it to achieve NSP, facilitating the maintenance of long-distance relationships between texts. The BERT was pre-trained on English Wikipedia (2.5B words) and BookCorpus (800M words). BERT has two models. The BERT_BASE has 12 layers, 768 widths, 12 heads, and a total of 110M parameters. The BERT_LARGE has 24 layers, 1024 widths, 16 heads, and a total of 340M parameters.

**GPT-2**[46]: Generative Pretrained Transformer 2 (GPT2) was developed by OpenAI researchers in 2019. GPT-2 models language using transformer decoders. GPT-2 is designed for predicting the next sentence in sentences. It achieves this using the architecture of GPT-1 with an extra normalization layer to the input of each sub-block and after the final self-attention layer. GPT-2's input includes texts' weight embeddings and their positional embeddings for context extraction. The input is then passed through the multi-head attention layer in the transformer decoder blocks, followed by a feed-forward layer. Finally, the softmax outputs a probability distribution. The GPT-2 was trained on an extensive corpus of WebText (40 GB of text, 8 million documents, from 45 million webpages upvoted on Reddit). It has 48 layers, 1600 widths, and 1.5B parameters.

**BI-LSTM-CRF**[44]: The Bidirectional LSTM CRF network combines a bidirectional LSTM network with a CRF network. It efficiently makes use of both long-distance past and future input features via the bidirectional LSTM layer as well as sentence-level tag information via the CRF layer. The initial layer of the model, designed to capture the semantics of the input text sequence, is a Bi-LSTM network. This layer's output is then fed into a CRF layer, which generates a probability distribution over the tag sequence by utilizing the interdependencies among the labels of the entire sequence.

**GPT-3**[47]: The Generative Pretrained Transformer 3 (GPT-3) employs a similar model architecture as GPT-2; however, it incorporates both alternating dense and locally banded sparse attention patterns in the transformer layers. The latest GPT-3 with 175B parameters was trained on the Common Crawl Dataset, expanded WebText Dataset, two internet-based books corporas, and English Wikipedia, using a context window of 2048 tokens. The pre-training steps mirror those of GPT-2, relatively but with increased dataset scale, data diversity, and training duration. The checkpoint used for experiments in this paper is "gpt-3.5-turbo-0613".

**GPT-4**[48]: The Generative Pretrained Transformer 4 (GPT-4) was released by OpenAI in 2023 without disclosure of model architecture, training, dataset, etc. Different from GPT-3, GPT-4 was fine-tuned using Reinforcement Learning from Human Feedback (RLHF). The checkpoint used for experiments in this paper is GPT-4-0613.

**PaLM 2**[49]: The Pathways Language Model (PaLM) 2 is built upon the Transformer. Specific information regarding its model size and architecture has not been shared in public. From available information, however, the predecessor of PaLM 2, known as PaLM[50], has 540B parameters, and deploys a standard Transformer architecture in a decoder-only configuration, albeit with certain alterations (e.g., SwiGLU Activation, RoPE embedding, etc). The checkpoint used for experiments in this paper is "bison-001", the second largest model in the PaLM 2 family.